\documentclass[conference]{IEEEtran}
\IEEEoverridecommandlockouts
\usepackage{multirow}
\usepackage{cite}
\usepackage{amsmath,amssymb,amsfonts}
\usepackage{algorithmic}
\usepackage{graphicx}
\usepackage{textcomp}
\usepackage{xcolor}
\usepackage{subcaption}
\usepackage{enumitem}

\def\BibTeX{{\rm B\kern-.05em{\sc i\kern-.025em b}\kern-.08em
    T\kern-.1667em\lower.7ex\hbox{E}\kern-.125emX}}
\begin{document}

\title{Intent-Driven Smart Manufacturing Integrating Knowledge Graphs and Large Language Models
%\title{Employing Knowledge Graphs and Large Language Models for Smart Manufacturing
%{\footnotesize \textsuperscript{*}Note: Sub-titles are not captured for https://ieeexplore.ieee.org  and
%should not be used}

\thanks{Identify applicable funding agency here. If none, delete this.}
}

\author{
\IEEEauthorblockN{Takoua Jradi\textsuperscript{1}, John Violos\textsuperscript{1}, Dimitrios Spatharakis \textsuperscript{2}, Lydia Mavraidi \textsuperscript{2}, Ioannis Dimolitsas \textsuperscript{2}, Aris Leivadeas\textsuperscript{1}, \\Symeon Papavassiliou \textsuperscript{2}}
\IEEEauthorblockA{\textsuperscript{1}Department of Software and IT Engineering, École de technologie supérieure, Montreal, Canada \\
\textsuperscript{2}School of Electrical \& Computer Engineering, National Technical University of Athens }}

\maketitle

\begin{abstract}
\begin{abstract}
The increasing complexity of smart manufacturing environments demands interfaces that can translate high-level human intents into machine-executable actions. This paper presents a unified framework that integrates instruction-tuned Large Language Models (LLMs) with ontology-aligned Knowledge Graphs (KGs) to enable intent-driven interaction in Manufacturing-as-a-Service (MaaS) ecosystems. We fine-tune Mistral-7B-Instruct-V02 on a domain-specific dataset, enabling the translation of natural language intents into structured JSON requirement models. These models are semantically mapped to a Neo4j-based knowledge graph grounded in the ISA-95 standard, ensuring operational alignment with manufacturing processes, resources, and constraints. Our experimental results demonstrate significant performance gains over zero-shot and 3-shots baselines, achieving 89.33\% exact match accuracy and 97.27\% overall accuracy. This work lays the foundation for scalable, explainable, and adaptive human-machine collaboration in smart manufacturing, with promising implications for real-time applications.
\end{abstract}

\end{abstract}


\section{Introduction}

The manufacturing industry is rapidly evolving toward intelligent, autonomous, and human-centric paradigms under the banner of Industry 4.0. At the heart of this transformation lies Smart Manufacturing (SM), which integrates advanced technologies such as Cyber-Physical Systems (CPS), the Industrial Internet of Things (IIoT), Artificial Intelligence (AI), and Knowledge Engineering to optimize decision-making, reduce downtime, and enable real-time adaptability \cite{lee_cyber-physical_2015}. This new paradigm is not merely about automation, it seeks to embed contextual intelligence and flexible responsiveness into production environments.

Central to this transformation is the integration of CPS, which facilitates real-time feedback loops between physical assets and their digital counterparts. As highlighted by Wang et al. \cite{wang_current_2015}, CPS enables tighter coupling between production processes and control logic, paving the way for agile manufacturing workflows. However, with increasing complexity and heterogeneity in both industrial assets and operational goals, manufacturers face growing challenges in semantic interoperability, decision traceability, and user-centric orchestration.

To tackle these challenges, the paradigm of servitization, the transformation from product-centric to service-oriented manufacturing, has gained traction. This paradigm emphasizes delivering value through services tailored to individual customer requirements, often under the umbrella of Manufacturing-as-a-Service (MaaS) \cite{baines_servitization_2009}. In such settings, expressing desired outcomes at a high level becomes essential. This has led to the rise of intent-driven manufacturing, where users articulate goals or intents in natural language, and intelligent systems autonomously interpret and fulfill them.

Intent-driven approaches provide a powerful abstraction layer, enabling stakeholders to express operational objectives without needing to define low-level execution details. For example, a factory operator might specify: “Reduce energy consumption of Line B without impacting throughput,” leaving it to the system to reconfigure machine parameters or reschedule tasks. As discussed by Shrouf et al. \cite{shrouf_smart_2014}, such intent-aware systems are especially suited to enhancing servitization strategies, allowing production to align more fluidly with customer and enterprise objectives.
However, realizing this vision poses significant technical hurdles. The inherent ambiguity, variability, and contextual dependency of natural language make intent interpretation a non-trivial task.

Traditional rule-based systems lack the flexibility and semantic depth to accurately parse and act upon open-ended user intents. To bridge this gap, recent advances in Knowledge Graphs (KGs) and Large Language Models (LLMs) present a promising avenue\cite{yang_knowledge_2024}.

Knowledge Graphs provide a structured, machine-interpretable representation of domain knowledge encompassing entities like machines, processes, materials, and their interrelationships. They enable systems to reason about operational constraints, dependencies, and histories. On the other hand, LLMs such as GPT, LLaMA, and Mistral excel at understanding and generating natural language, making them ideal for capturing user intents and generating semantically meaningful responses \cite{reitemeyer_applying_2025}.

Despite their individual strengths, current efforts often treat KGs and LLMs as separate tools, missing the opportunity for synergistic integration.

In this paper, we present a unified framework for intent-driven smart manufacturing that leverages instruction-tuned large language models to translate high-level natural language intents into structured, machine-executable requirement models. Rather than merely identifying goals, our approach operationalizes user intent by generating precise JSON-based specifications that align with domain-specific manufacturing processes. This enables automation systems to reason over structured requirements instead of ambiguous text, promoting transparency, reusability, and decision support.

While the current system relies solely on LLM-based reasoning, it lays the groundwork for future integration with manufacturing knowledge graphs (KGs) and domain ontologies. This will allow the system to dynamically map structured intents onto semantically relevant subgraphs, enabling context-aware reasoning and more adaptive behavior. Ultimately, our vision is to build an intent interface where human operators think in high-level goals, and smart manufacturing systems respond with structured, explainable, and operationally grounded actions.

The rest of this paper is organized as follows: Section~\ref{sec:related_work} reviews related approaches; Section~\ref{sec:proposed_model} describes our work; Section~\ref{sec:experiments} presents experimental evaluations; and Section~\ref{sec:conclusion} outlines future research directions.

\section{Related Work}
\label{sec:related_work}
Recent advances in knowledge graphs (KGs) and large language models (LLMs) have enabled new approaches to intent understanding and automation in smart manufacturing. Several studies have explored ontology-based KG frameworks to structure manufacturing knowledge. Kang et al.~\cite{kang_ontology_2024} proposed AMPO, an ontology for aircraft maintenance, using BERT-based models for entity and relation extraction. However, their approach requires manual validation and lacks dynamic intent interpretation. Similarly, Shima et al.~\cite{shima_omega_2025} introduced OmEGa, an ontology-guided LLM framework for manufacturing KGs, but it depends heavily on predefined prompts and struggles with ambiguous intents.

LLMs have also been increasingly applied to KG tasks, such as entity extraction and link prediction. Hou et al.~\cite{hou_low-resource_2025} used LLMs for knowledge distillation in low-resource KGs, employing rethink and open prompts to enhance reasoning. However, their method does not address real-time intent processing. The integration of LLMs with KGs for knowledge completion and text-to-entity mapping has also gained attention. Yao et al.~\cite{yao_exploring_2025} proposed KG-LLM, a framework that treats KG triples as text sequences and fine-tunes smaller LLMs (e.g., LLaMA-7B, ChatGLM-6B) to achieve state-of-the-art performance in tasks like triple classification and relation prediction. Their work demonstrates that instruction tuning can effectively bridge the gap between pre-trained LLMs and KG structures, outperforming larger models like ChatGPT and GPT-4 in specific tasks.Similarly, Kartsaklis et al.~\cite{kartsaklis_mapping_2018} introduced a Multi-Sense LSTM (MS-LSTM) for mapping natural language text to KG entities. Their approach leverages textual features to enhance the entity space and employs a dynamic disambiguation mechanism to handle polysemy. This method achieves high accuracy in text-to-entity mapping and reverse dictionary tasks, showcasing the potential of combining neural architectures with KG embeddings for semantic understanding.

In the networking domain, the KARMA framework by Lu and Wang~\cite{lu_karma_2025} enables the transformation of multi-modal inputs into semantic representations using KGs and multi-agent LLMs. KARMA supports automatic schema alignment and ontology-guided enrichment, but it is not specifically designed for intent interpretation in manufacturing scenarios or for managing domain-specific execution constraints.

Wang et al.~\cite{wang_intent-based_2023} explore a holistic approach to intent-based network management by integrating KGs and intent modeling. Their architecture employs layered reasoning over domain ontologies to align user intents with network configurations. While effective in vertical industries, their model relies on static templates and lacks mechanisms to dynamically adapt to complex, evolving intents.

\textbf{Our work advances beyond these limitations by introducing a unified framework that combines instruction-tuned LLMs with a structured knowledge graph to support intent-driven smart manufacturing.} Unlike prior approaches that rely solely on static ontologies or prompt-based methods, we fine-tune a general-purpose LLM to generate structured requirement models, which are then dynamically mapped onto a custom Neo4j-based KG grounded in the ISA-95 standard.
Our ontology formalizes manufacturing processes, resources, and constraints, enabling semantic reasoning and subgraph extraction based on user intent. This KG-LM integration bridges unstructured user input with executable manufacturing knowledge, offering both interpretability and operational flexibility across distributed MaaS environments.
\section{Proposed Model}
\label{sec:proposed_model}

\subsection{Overview}
 We propose a unified \textbf{intent translation model} that converts natural language intents into formalized \textbf{knowledge graph (KG)} representations through a structured, multi-stage pipeline. This pipeline integrates \textbf{large language models (LLMs)}, \textbf{semantic requirement modeling}, and \textbf{domain ontologies} to ensure interpretable and operationally executable outputs.

\begin{figure}[h]
    \centering
    \includegraphics[width=1\linewidth]{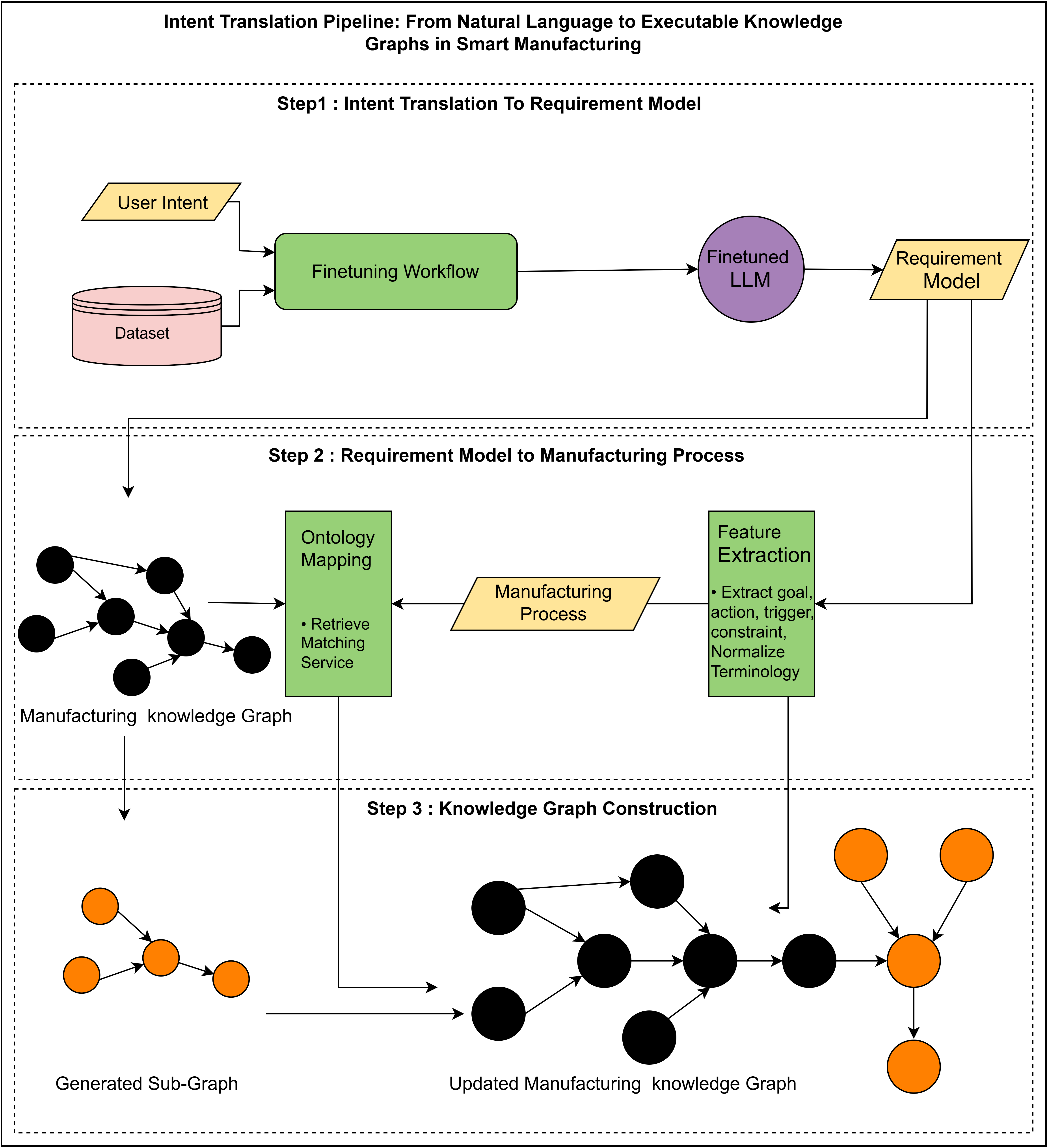}
    \caption{Intent Translation Pipeline: from natural language input to structured requirement model.}
    \label{fig:intent-model}
\end{figure}

As illustrated in Figure~\ref{fig:intent-model}, the process consists of three main stages:

\begin{itemize}
    \item \textbf{Step 1: Intent Translation to Requirement Model} A user expresses an operational goal in natural language. The system processes the user intent through an instruction-based prompt using a \textbf{fine-tuned LLM}, which generates a structured \texttt{requirement model} in JSON format. This model specifies the goal, trigger, actions, and associated constraints required for execution.

    \item \textbf{Step 2: Requirement Model to Manufacturing Process}  The generated requirement model undergoes \textit{feature extraction} to identify semantic components (e.g., goals, actions, constraints). These components are aligned with a manufacturing process ontology using a \textit{retrieve-and-match} mechanism, enabling domain-specific interpretation and operational alignment.

    \item \textbf{Step 3: Knowledge Graph Construction} The resulting process model is translated into a knowledge subgraph and merged with the broader manufacturing knowledge graph, incrementally updating the system’s operational understanding.
\end{itemize}

\subsection{User Intents}
 User intents serve as high-level operational goals expressed in natural language. These intents typically convey both functional requirements (e.g., scheduling updates, triggering actions) and non-functional requirements (e.g., accuracy thresholds, latency limits, resource constraints), enabling operators to specify system behavior in a human-readable format.

\paragraph{Illustrative Example}

An example of a user intent extracted from real-world fleet management operations is shown below:

\begin{quote}
\textit{"Automatically update the internal fleet schedule within 5 seconds. All applied changes must achieve at least 99.9\% accuracy. The system shall ensure 100\% data integrity, maintain 99.99\% uptime availability, and perform updates using no more than 65\% of CPU and memory resources."}
\end{quote}
\subsection{Structured Output Format}

To enable machine interpretability, each intent is translated into a structured requirement model in JSON format. The semantic fields extracted include \texttt{goal}, \texttt{mode}, \texttt{trigger}, and \texttt{action}, where the \texttt{action} object also defines detailed \texttt{constraint} specifications. For the intent above, the translated requirement model is:

\begin{verbatim}
{
  "goal": "UpdateInternalFleetSchedule",
  "mode": "automated",
  "trigger": {
    "condition": "FleetChangeDetected"
  },
  "action": {
    "type": "ApplyScheduleUpdate",
    "constraint": {
      "timeLimit": "5 seconds",
      "accuracy": ">=99.9%",
      "dataIntegrity": "100%",
      "availability": ">=99.99%",
      "resourceUtilization": {
        "CPU": "<=65%",
        "Memory": "<=65%"
      }
    }
  }
}
\end{verbatim}
\begin{figure*}[t]
    \centering
    \begin{subfigure}[b]{0.49\textwidth}
        \includegraphics[width=\linewidth]{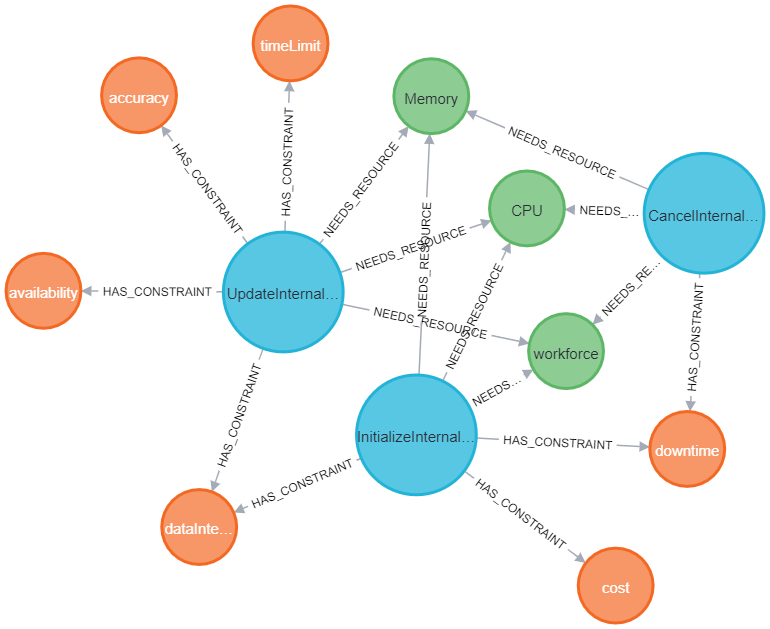}
        \caption{Instance of the KG for Aircraft Maintenance.} % Replace with appropriate caption
        \label{fig:ontology}
    \end{subfigure}\hfill
    \begin{subfigure}[b]{0.49\textwidth}
        \includegraphics[width=\linewidth]{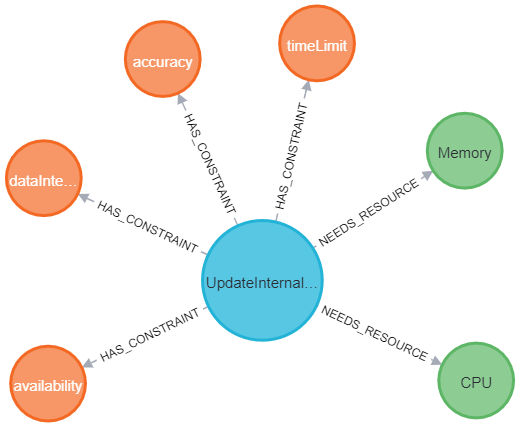} % Ensure extension is correct
        \caption{Extracted subgraph of the KG.}
        \label{fig:extracted_kg}
    \end{subfigure}
    \caption{KG retrieval of information according to the specified user's intent.}
    \label{fig:kg}
\end{figure*}

\subsection{Knowledge Graph}
The KGs offer a structured way of representing information about the operations of the manufacturing domain. Several papers in the bibliography organize this information into different data schemas, trying to unify the information between different manufacturing sectors \cite{buchgeher2021knowledge}. To this extent, several ontologies have been introduced to achieve this by incorporating information coming from the manufacturing operations, tailored to the needs of the specific use cases. 

A seminal work in this field is the MASON ontology \cite{1633441} which is based on the Web Ontology Language (OWL) to provide abstractions for various manufacturing systems. Several other Standardization efforts have been done, including Enterprise-Control System Integration (ISA-95), Basic Formal Ontology, and ADAptive holonic COntrol aRchitecture for distributed manufacturing systems. Recently, the Asset Administration Shell (AAS) \cite{bader2019semantic} has been introduced to provide a digital representation of manufacturing systems, allowing the creation of Digital Twins and enabling the digital transformation of the manufacturing domain. 

In this paper, we introduce a novel ontology specifically designed in alignment with the ISA-95 standard, which serves as a foundational framework for integrating manufacturing systems. The proposed ontology has the following core entities:
\begin{itemize}
    \item \textbf{Manufacturing Processes (MP):}
    These refer to specific operations or tasks carried out within a manufacturing setting. 
    In the proposed ontology, the MPs act as the main foundational block.
    \item \textbf{Manufacturing Resources (MR):} 
    These include various assets, tools, and input elements necessary to support and execute manufacturing processes effectively. For example, the MRs may represent raw materials, production equipment, human operators, and computational resources like CPU and memory. This broad definition ensures the coverage of various potential dependencies of manufacturing operations.
    \item \textbf{Process Constraints (PCs):} Each MP is associated with a set of entities that declare the parameters, constraints, and Key Performance Indicators (KPIs) that define its operational behavior, performance expectations, quality criteria, and other relevant process-specific requirements.  
\end{itemize}
The proposed ontology declares mappings between each MP and one or more MRs, thereby modeling the dependencies necessary for the execution of manufacturing tasks. Moreover, each MP is associated with PCs to declare the operational constraints. Therefore, the ontology enables the accurate representation of manufacturing operations in the context of MaaS, paving the way for improved interoperability and semantic integration across distributed manufacturing environments.

To accommodate each user-declared intent, the requirement model extracted by the Intent Translation Model must be mapped to a subgraph of the ontology that describes a specific manufacturing operation. Specifically, in this work, we develop the proposed ontology as a Neo4j KG instance. Then, to extract the respective subgraph of the KG, we query the KG instance according to the translated requirement model. This operation allows for (a) retrieving the corresponding entities and relationships between them and (b) updating the KG attributes of these relationships.

In Fig. \ref{fig:ontology}, an instance of the ontology is depicted describing the relationship between the entities for the manufacturing processes associated with Aircraft Maintenance as a Service domain. While the instance contains multiple MPs, MRs, and PCs relevant to various operations for this domain, in Fig. \ref{fig:extracted_kg} the corresponding subgraph for the given intent regarding the \textit{UpdateInternalFleetSchedule} MP is depicted, containing only the user-relevant information and the associated entities. To extract the subgraph, we utilize Cypher to query the Neo4j KG database. Finally, the relationships between the entities of the \textit{UpdateInternalFleetSchedule} are updated according to the translated intent to represent the new required values. 

\subsection{Large Language Model}
Large Language Models (LLMs) are transformer-based neural networks trained on large-scale text corpora to perform tasks such as generation, summarization, and instruction following. They are foundational to translating natural language into structured representations.

For our task, we use Mistral 7B Instruct-V2, a lightweight, instruction-tuned, decoder-only model optimized for structured output and efficient fine-tuning.
\paragraph{Why Mistral 7B Instruct?}
We adopt Mistral 7B Instruct for its architectural efficiency, structured output reliability, and compatibility with resource-constrained fine-tuning. Unlike LLaMA or T5, Mistral incorporates Sliding Window Attention (SWA) and Grouped Query Attention (GQA) to support long-context understanding and reduce compute cost, enabling it to rival larger models like LLaMA 13B while using just 7B parameters. Its instruction tuning produces more deterministic and format-consistent completions, which is essential for structured JSON generation. Mistral is also highly compatible with 4-bit quantization methods such as QLoRA, delivering stable performance without requiring extensive calibration—unlike LLaMA or T5, which are either less adaptable or not optimized for causal decoding. Its compact size and strong decoding behavior make it ideal for training and inference on limited hardware. Furthermore, Mistral is open-source and well-supported within ecosystems like Hugging Face, PEFT, and llama.cpp, ensuring ease of deployment, fine-tuning, and long-term maintainability.

\section{Experimental Evaluation}
\label{sec:experiments}
\subsection{Simulation Setting}
All experiments were executed on a high-performance Linux workstation equipped with an Intel(R) Core(TM) i9-14900 CPU @5.8 GHz, alongside 125 GiB of RAM. For accelerated training, we utilized an NVIDIA RTX 6000 Ada Generation GPU with support for tensor cores and mixed precision computation, making it well-suited for 4-bit quantized large language model optimization.
\subsection{Dataset Preparation}
To enable effective training and evaluation of our intent translation model in the context of smart manufacturing, we curated a structured dataset comprising \textbf{2,580 annotated samples}. Each sample pairs a high-level user intent expressed in natural language with its corresponding structured requirement model formatted in JSON.

Due to the scarcity of publicly available datasets in the field of \textit{intent-to-requirement translation}, particularly within manufacturing environments, we constructed this dataset from scratch. This task is inherently challenging given that most existing intent-based datasets lack the domain specificity, structural complexity, and constraint formalism required in smart manufacturing. As this line of research is still in its early stages, \textbf{no standard benchmark existed to meet our needs}.

To address this gap, we manually created a corpus of \textbf{2,850 domain-relevant raw examples} equally distributed across \textbf{three representative manufacturing processes}, selected for their industrial relevance and operational diversity:
\begin{itemize}
\item \textbf{UpdateInternalFleetSchedule:}\\
This category includes user intents related to the dynamic adjustment of internal fleet schedules in response to changes in operational status or production demands.  
\textit{Associated constraints:} \texttt{timeLimit}, \texttt{accuracy}, \texttt{dataIntegrity}, \texttt{availability}, and \texttt{resourceUtilization} \texttt{(CPU, Memory)}.

\item \textbf{RequestEmptyContainers:}\\
This category focuses on intents involving the request, reservation, or redistribution of empty containers across logistics nodes, with a strong emphasis on timeliness and resource availability.  
\textit{Associated constraints:} \texttt{responseTime}, \texttt{containerAvailability}, \texttt{deliveryDeadline}, \texttt{priorityLevel}, and \texttt{resourceEfficiency}.

\item \textbf{DynamicContainerRouteOptimization :}\\
This category captures intents aimed at generating optimized routing plans for container movements under operational constraints. The emphasis is on improving system efficiency while adhering to logistical and performance thresholds.  
\textit{Associated constraints:} \texttt{latency}, \texttt{optimizationAccuracy}, \texttt{fuelReduction}, \texttt{throughput}, and \texttt{efficiencyGain}.
\end{itemize}
 To simulate a diverse range of linguistic expressions while preserving domain semantics, we adopted a hybrid data generation strategy. First, a set of base templates was manually crafted to reflect typical operational intents. These templates were then expanded using AI-powered paraphrasing tools such as ChatGPT to generate syntactically varied yet semantically consistent reformulations. This approach ensured the inclusion of varied phrasing, passive and active voice constructions, and realistic variations in terminology.
Each generated intent was mapped to a structured requirement model manually designed to follow a rigid JSON schema that includes fields such as \texttt{goal}, \texttt{mode}, \texttt{trigger}, \texttt{action}, and quantifiable \texttt{constraints}. 
Human-in-the-loop validation was employed to ensure data quality. All generated samples were reviewed by at least one human annotator familiar with industrial operations and constraint-based modeling. 
\subsection{Model Implementation}
\begin{figure*}[htbp]
    \centering
    \includegraphics[width=\textwidth]{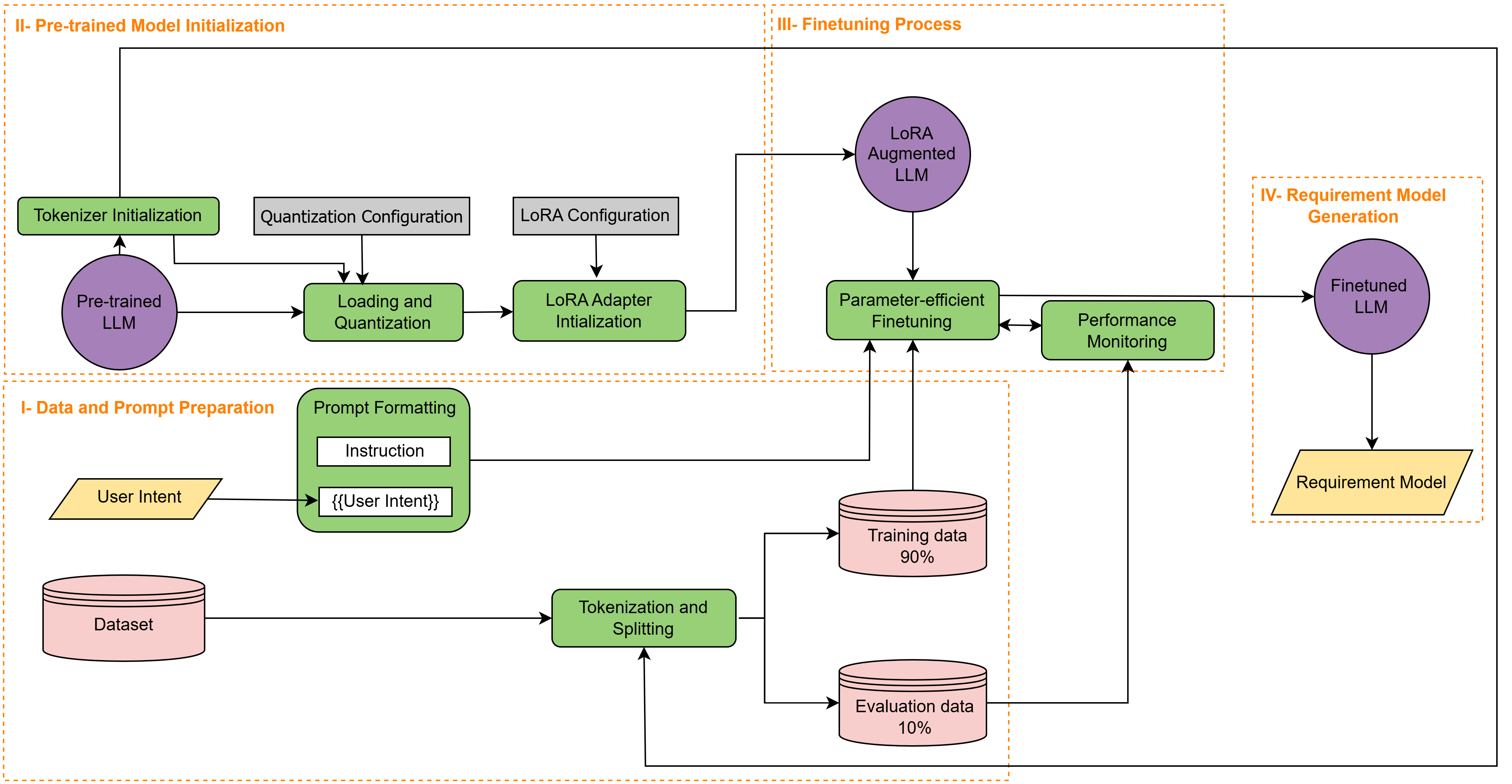}
    \caption{Finetuning Workflow.}
    \label{fig:finetuning}
\end{figure*}
We employ the \textbf{Mistral-7B-Instruct-v0.2} model as the backbone for translating natural language intents into structured requirement models. The model is fine-tuned using a supervised instruction-following approach, specifically optimized for the task of intent-to-requirement generation.

To ensure training efficiency and compatibility with consumer-grade hardware, we apply 4-bit quantization using the BitsAndBytes library with the following configuration:
\begin{itemize}
    \item \texttt{bnb\_4bit\_compute\_dtype = torch.float16}
    \item \texttt{bnb\_4bit\_quant\_type = "nf4"}
    \item \texttt{bnb\_4bit\_use\_double\_quant = True}
\end{itemize}

We use the Hugging Face Transformers library for model loading and training. Tokenization is performed using the official tokenizer associated with Mistral to ensure consistency with pre-trained embeddings. We also resize the token embeddings to accommodate additional special tokens used for prompting.

To enable efficient adaptation, we integrate \textbf{Low-Rank Adaptation (LoRA)} using the PEFT framework. LoRA adapters are injected into the model’s attention layers, allowing only a small number of trainable parameters while keeping the base model frozen.

The fine-tuning process follows an instruction-based prompting strategy:
\begin{verbatim}
\begin{verbatim}
<s>[INST] Translate the following intent 
into a requirement model:
<user_intent> [/INST] <structured_output>
\end{verbatim}
This format aligns with Mistral's instruct-style training and ensures predictable, structured output in JSON format.

To optimize training time and GPU memory usage, we also enable \textbf{gradient checkpointing}. 
Figure~\ref{fig:finetuning} illustrates our full pipeline, consisting of the following stages:

\begin{enumerate}[label=\textbf{\Roman*.}, leftmargin=2em]
    \item \textbf{Data and Prompt Preparation:}  
    We construct a dataset of user intents paired with corresponding requirement models in JSON. Each sample is framed using an instruction template and tokenized. The dataset is split into 90\% training and 10\% evaluation for real-time performance monitoring.

    \item \textbf{Pre-trained Model Initialization:}  
    The Mistral-7B-Instruct-v0.2 model is initialized with 4-bit quantization using the BitsAndBytes library. Token embeddings are resized to support additional tokens. LoRA adapters are injected into attention layers, and gradient checkpointing is enabled for memory optimization.

    \item \textbf{Fine-Tuning Process:}  
    Fine-tuning is performed using the instruction-style dataset. Only LoRA parameters are updated, while the base model remains frozen. Validation is performed after each epoch using metrics such as loss, exact match accuracy, and F1-score.

    \item \textbf{Requirement Model Generation:}  
    At inference time, user intents are formatted using the same prompt template and passed to the fine-tuned model. The output is a structured JSON requirement model ready for integration into downstream scheduling, monitoring, or automation systems.
\end{enumerate}

\subsection{Training Setup \& Outcomes}

The model was trained using the Hugging Face Trainer API with the following hyperparameters:

\begin{itemize}
  \item Learning rate: \( 2 \times 10^{-4} \)
  \item Epochs: 10
  \item Batch size: 4
  \item Evaluation strategy: per steps
  \item Logging and saving every 100 steps
\end{itemize}
\begin{table}[h]
\centering
\caption{Performance of Fine-Tuned Model Across Epochs}
\label{tab:epoch_progress}
\begin{tabular}{|c|c|c|c|}
\hline
\textbf{Epochs} & \textbf{Training Loss} & \textbf{Validation Loss} & \textbf{Accuracy} \\
\hline
\textbf{1} & 0.078 & 0.071 & 96,57\% \\
\hline
\textbf{2} & 0.067 & 0.070 & 96.66\% \\
\hline
\textbf{3} & 0.067 & 0.063 & 96.71\% \\
\hline
\textbf{4} & 0.063 & 0.0612 & 96.79\% \\
\hline
\textbf{5} & 0.062 & 0.0610 &  96.86\% \\
\hline
\textbf{6} & \textbf{0.0623} & \textbf{0.0611} & \textbf{97.29\%} \\
\hline
\textbf{7} & 0.0612 & 0.0611 & 96.51\% \\
\hline
\textbf{8} & 0.0618 & 0.0612 & 96.14\% \\
\hline
\textbf{9} & 0.0618 & 0.0616 & 96.14\% \\
\hline
\textbf{10} & 0.0616 & 0.0615 & 96.00\% \\
\hline
\end{tabular}
\end{table}

The fine-tuned \textbf{Mistral-7B-Instruct-v0.2} model demonstrated strong performance in translating user intents into structured JSON requirement models.
The training progressed smoothly with consistent reductions in both training and validation loss across the initial epochs. Accuracy steadily improved, with the model achieving its highest accuracy of \textbf{97.29\%} at the \textbf{6th epoch}.After the sixth epoch, although the training loss continued to decline, validation loss plateaued and accuracy began to drop slightly. This divergence indicated the early onset of overfitting, highlighting the sixth epoch as the optimal stopping point.The model maintained strong generalization throughout training. No significant performance degradation was observed on the validation set, suggesting that the model remained stable and well-calibrated under the defined hyperparameters.
\subsection{Baselines Comparaison}

To assess the effectiveness of this approach, we conduct a comparative evaluation against three baselines:

\begin{enumerate}
\item \textbf{Zero-Shot Baseline:} Using the pre-trained Mistral-7B-Instruct-V02 model directly without any fine-tuning.
\item \textbf{Few-Shot Baseline:} Prompting the pre-trained model with a few in-context examples.
\item \textbf{Supervised Fine-Tuned Model (ours):} The Mistral-7B-Instruct-V02 model fine-tuned on 2,580 domain-specific intent-requirement pairs across three key manufacturing processes.
\end{enumerate}
This comparison allows us to quantify the gains from domain adaptation and instruction tuning. We evaluate each model using structured prediction metrics, including JSON validity, Exact Match Accuracy, and goal-specific constraint extraction performance (precision, recall, F1-score). The experimental setup and results are presented in the following sections.
\begin{table}[h]
\centering
\caption{Comparative Evaluation Metrics for Different Model Variants}
\label{tab:model_comparison_transposed}
\begin{tabular}{|l|c|c|c|}
\hline
\textbf{Metric} & \textbf{Zero-Shot} & \textbf{3-Shots} & \textbf{Fine-Tuned (Ours)} \\
\hline
Exact Match Accuracy & 0.00\% & 18.41\% & \textbf{89.33\%} \\
\hline
JSON Validity        & 0.00\% & 71.23\% & \textbf{100\%} \\
\hline
Precision            & 0.00\% & 36.72\% & \textbf{95.34\%} \\
\hline
Recall               & 0.00\% & 32.12\% & \textbf{94.96\%} \\
\hline
F1 Score             & 0.00\% & 34.33\% & \textbf{95.07\%} \\
\hline
Overall Accuracy     & 0.00\% & 38.96\% & \textbf{97.29\%} \\
\hline
\end{tabular}
\end{table}
Our fine-tuned model significantly outperforms both the zero-shot and few-shot baselines across all metrics. These results highlight the critical role of supervised instruction tuning for domain-specific structured prediction in smart manufacturing.

\subsection{Discussion}
\begin{figure*}[htbp]
    \centering
    \begin{subfigure}[b]{0.5\textwidth}
        \includegraphics[width=\linewidth]{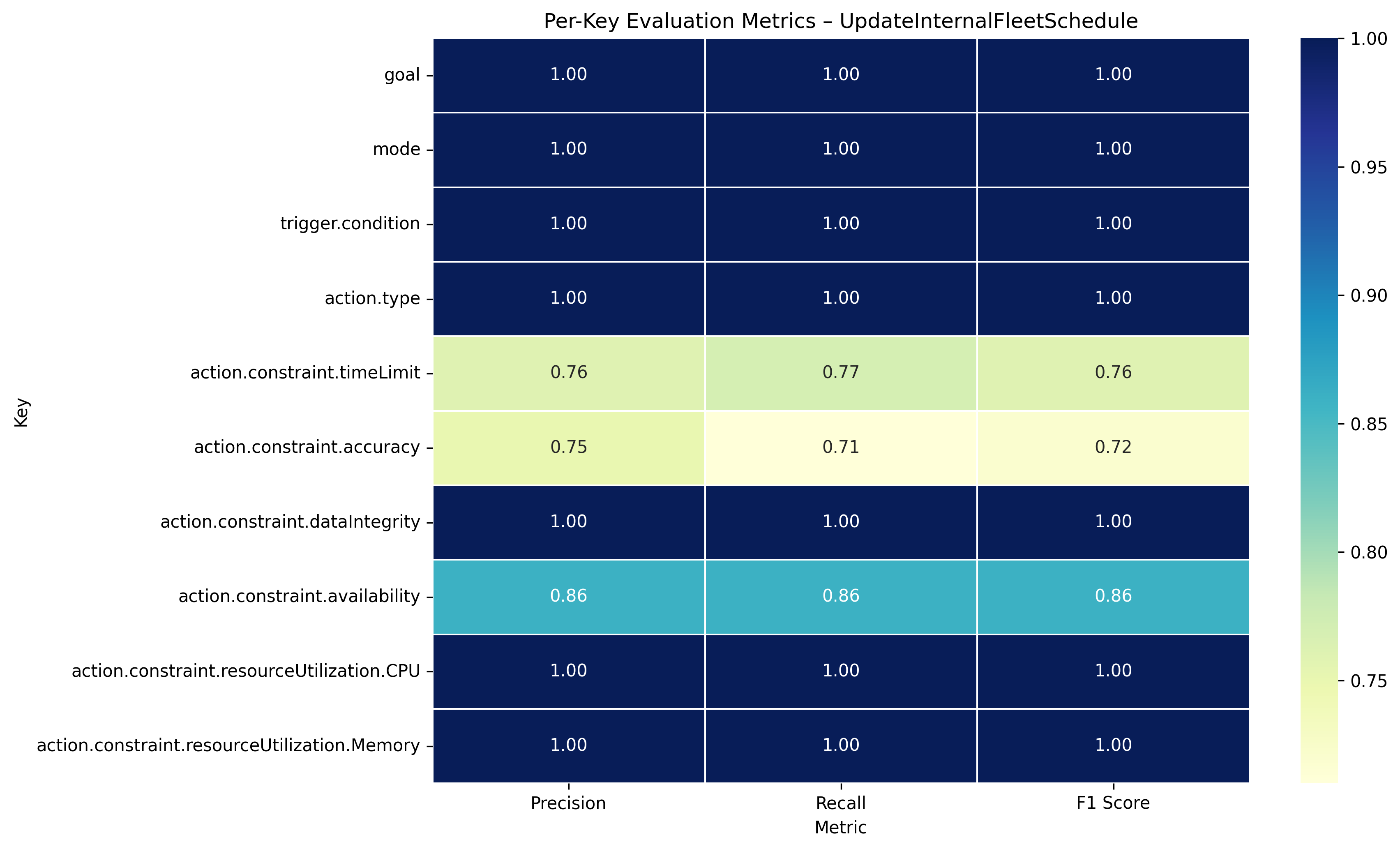}
        \caption{Evaluation Metrics for the "UpdateInternalFleetSchedule" Process.}        \label{fig:UpdateInternalFleetSchedule}
    \end{subfigure}\hfill
    \begin{subfigure}[b]{0.5\textwidth}
        \includegraphics[width=\linewidth]{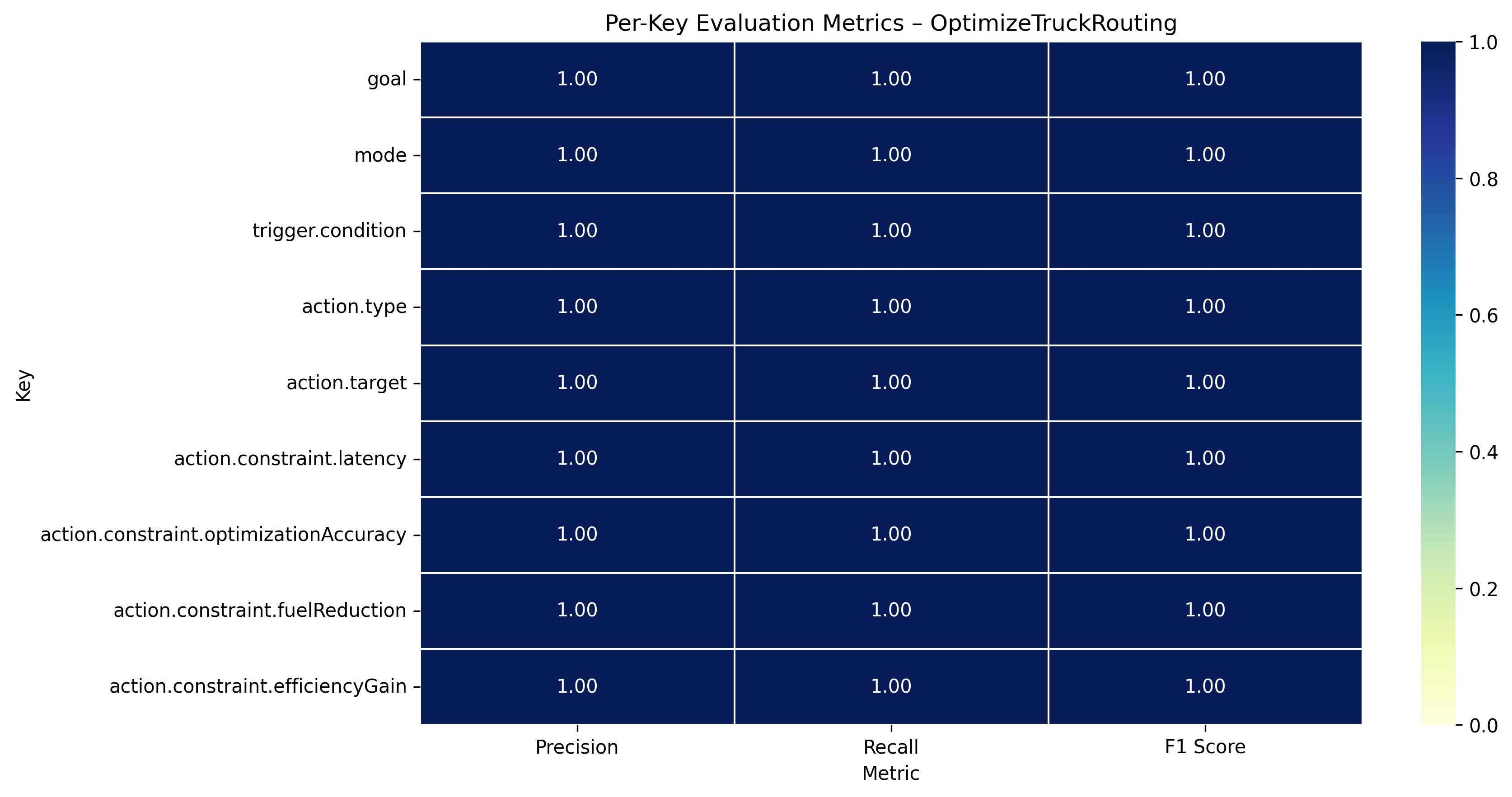} % Ensure extension is correct
        \caption{Evaluation Metrics for the "OptimizeTruckRouting" Process.}
        \label{fig:OptimizeTruckRouting}
    \end{subfigure}
     \begin{subfigure}[b]{0.5\textwidth}
        \includegraphics[width=\linewidth]{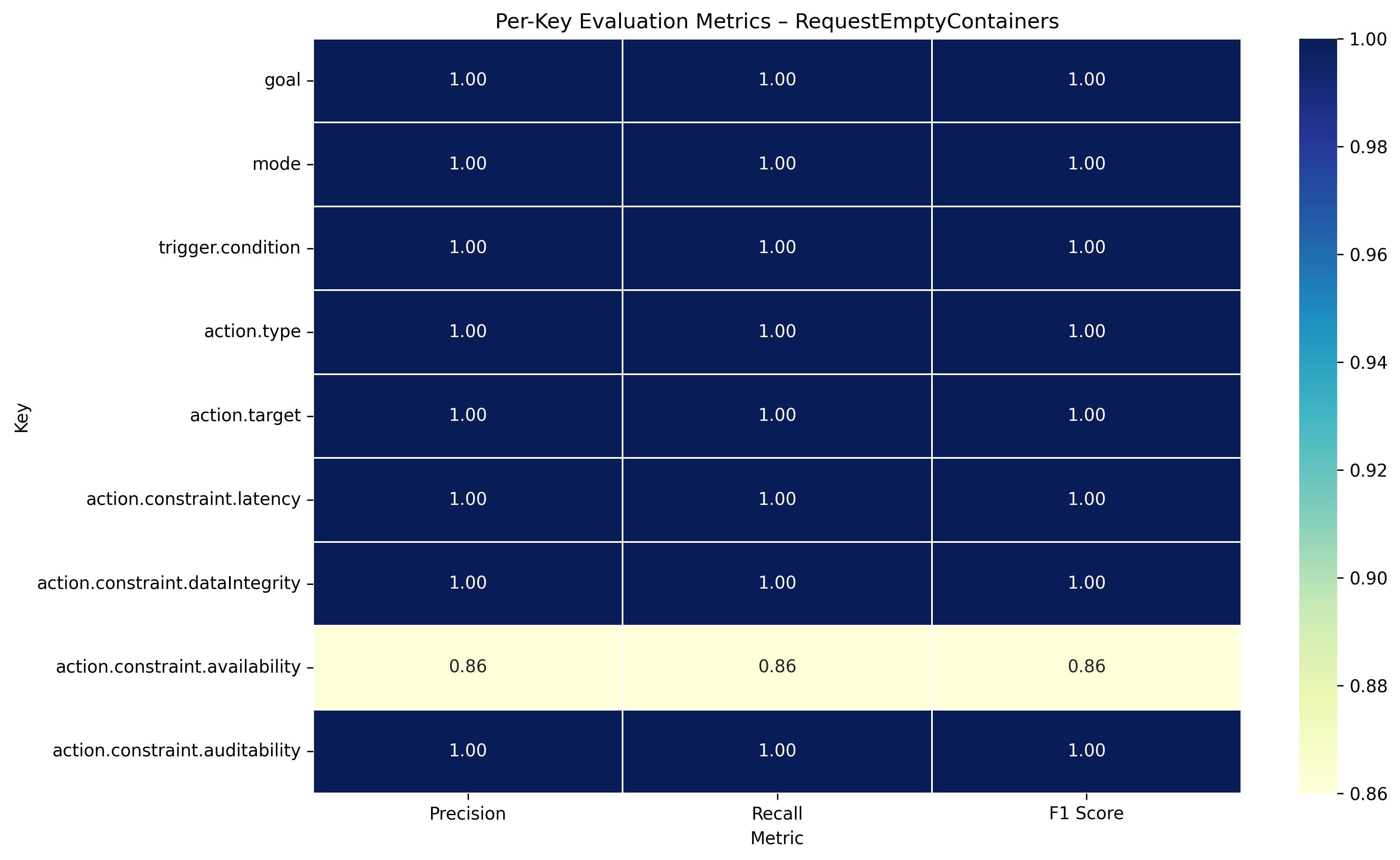}
        \caption{Evaluation Metrics for the "RequestEmptyContainers" Process.} % Replace with appropriate caption
        \label{fig:RequestEmptyContainers}
    \end{subfigure}\hfill
    \begin{subfigure}[b]{0.5\textwidth}
        \includegraphics[width=\linewidth]{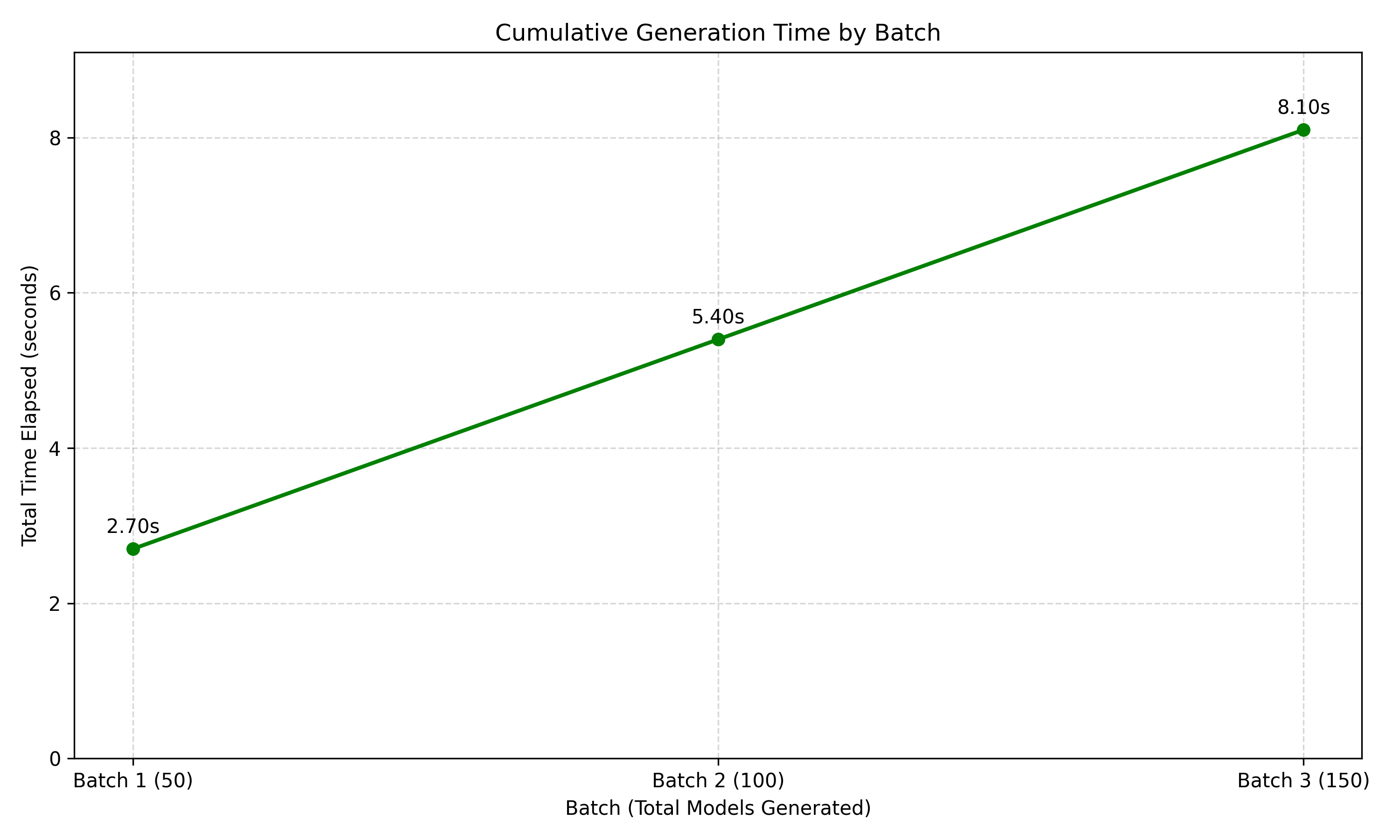} % Ensure extension is correct
        \caption{Cumulative Inference Time by Batch Size.}
        \label{fig:batch_time}
    \end{subfigure}\hfill
    \label{fig:kg}
\end{figure*}

To gain a detailed understanding of the model’s structured prediction capabilities, we analyze performance on a per-key basis across the three core manufacturing process. The evaluation focuses on precision, recall, and F1-score for each structured JSON field. For this purpose, we used a test dataset of 150 samples, 50 per process and prompted all models using the same instruction template to ensure consistency across evaluations. The results are visualized using heatmaps in Figures~\ref{fig:UpdateInternalFleetSchedule}, \ref{fig:OptimizeTruckRouting}, and \ref{fig:RequestEmptyContainers}.
\paragraph{UpdateInternalFleetSchedule.}
As shown in Figure~\ref{fig:UpdateInternalFleetSchedule}, the model performs with high consistency, achieving perfect scores (1.00) on all key fields except minor variations in constraints like time and availability. These deviations reflect the inherent complexity of interpreting such constraints, but the model still reliably captures the core intent structure.

\paragraph{OptimizeTruckRouting.}
Figure~\ref{fig:OptimizeTruckRouting} shows perfect performance across all fields. This result confirms the model’s ability to generalize effectively when intent phrasing is clear and consistently represented in training data.

\paragraph{RequestEmptyContainers.}
As illustrated in Figure~\ref{fig:RequestEmptyContainers}, the model achieves strong results, with only a slight dip (0.86) in extracting availability constraints. This is likely due to phrasing variability, yet overall output remains highly accurate and structurally sound.

\paragraph{Inference Efficiency.}
Figure~\ref{fig:batch_time} demonstrates that inference time scales linearly with batch size, averaging 54ms per sample. This confirms the model’s suitability for real-time deployment in industrial settings.

\section{Conclusion and Future Work}
\label{sec:conclusion}

This paper introduced a unified framework for intent-driven smart manufacturing that combines instruction-tuned language models with ontology-aligned knowledge graphs. By fine-tuning the Mistral-7B-Instruct-v0.2 model on a domain-specific dataset of 2,580 annotated samples, we enabled the automatic translation of high-level natural language intents into structured, machine-executable requirement models. These outputs are mapped to a Neo4j-based knowledge graph grounded in the ISA-95 standard, enabling semantic alignment with manufacturing processes, resources, and constraints.
Unlike prior approaches that rely solely on static prompts or pre-built ontologies, our method generates domain-compliant requirement structures that can be queried and manipulated using graph-based operations. This allows user intents to be interpreted in context, integrated into existing semantic models, and used to update manufacturing knowledge dynamically.

Looking ahead, we aim to generalize our framework across multiple manufacturing pilots to support cross-domain adaptability. We also plan to integrate Retrieval-Augmented Generation (RAG) to improve contextual relevance, particularly for unseen intents, and enable online fine-tuning to support continuous adaptation in dynamic environments. Together, these enhancements will transform the system into a scalable, flexible, and context-aware interface for intent-driven interaction in Manufacturing-as-a-Service (MaaS) ecosystems.

\section*{Acknowledgment}
This work was funded by the European Union’s Horizon Europe research and innovation program under grant agreement No. 101177842 Unified Modeling and Automated Scheduling for Manufacturing as a Service (UniMaaS)

\bibliographystyle{IEEEtran}
\bibliography{references}

\end{document}